\documentclass{article}

\usepackage{arxiv}

\usepackage[utf8]{inputenc} % allow utf-8 input
\usepackage[T1]{fontenc}    % use 8-bit T1 fonts
\usepackage{hyperref}       % hyperlinks
\usepackage{url}            % simple URL typesetting
\usepackage{booktabs}       % professional-quality tables
\usepackage{amsfonts}       % blackboard math symbols
\usepackage{nicefrac}       % compact symbols for 1/2, etc.
\usepackage{microtype}      % microtypography
\usepackage{lipsum}		% Can be removed after putting your text content
\usepackage{graphicx}
\usepackage{natbib}
\usepackage{doi}
\usepackage{soul}
\usepackage{caption}
\usepackage{subcaption}

\usepackage{titlesec}
\usepackage{titletoc} % Required for Table of Contents integration
\usepackage{chngcntr} % Required for counter management

\usepackage{xspace}
\usepackage{listings}
\usepackage{algorithm}
\usepackage{algpseudocode}
\usepackage[edges]{forest}
\forestset{
  mytree/.style={
    for tree={
      draw,
      rounded corners,
      align=left,
      inner sep=3pt,
      s sep=8pt,
      l sep=10pt,
      font=\sffamily\small,
      edge={-Latex},
    },
  },
  selected/.style={
    fill=blue!6,
    edge={ultra thick,blue},
    label=right:{\footnotesize\textbf{Selected for further evolution}}
  },
  pruned/.style={
    draw=gray!60,
    text=gray!70,
    edge={dashed,gray},
    label=right:{\footnotesize Pruned due to poor performance}
  }
}
\usepackage{tikz}
\usetikzlibrary{arrows.meta,positioning,fit,shapes.misc,calc}
\tikzset{
  box/.style={draw, rounded corners=2pt, inner sep=4pt, align=center},
  thinbox/.style={draw, rounded corners=1.5pt, inner sep=3pt, align=center, font=\footnotesize},
  flow/.style={-Latex, line width=0.7pt},
  dashedflow/.style={-Latex, line width=0.6pt, dashed},
  group/.style={draw, rounded corners=3pt, inner sep=6pt, densely dotted},
  title/.style={font=\bfseries},
}

\titleclass{\subsubsubsection}{straight}[\subsubsection]

\newcounter{subsubsubsection}
\counterwithin{subsubsubsection}{subsubsection}
\renewcommand{\thesubsubsubsection}{\thesubsubsection.\arabic{subsubsubsection}}

\titleformat{\subsubsubsection}
  {\normalfont\normalsize\bfseries} % Format: small, bold font
  {\thesubsubsubsection}       % Label: the defined counter format
  {0.5em}                      % Separation: 0.5em space between label and title
  {}                           % Before code: nothing here
  []                           % After code: nothing here

\titlespacing*{\subsubsubsection}
  {0pt}                        % Left margin (indentation)
  {2ex plus .1ex minus .1ex}   % Vertical space before title
  {1ex plus .1ex minus .1ex}   % Vertical space after title

\newcommand{\tool}{\textsc{PhyloEvolve}\xspace}
\usepackage{amsmath,amsfonts,bm}

\def\eqref#1{equation~\ref{#1}}
\def\1{\bm{1}}

\DeclareMathAlphabet{\mathsfit}{\encodingdefault}{\sfdefault}{m}{sl}
\SetMathAlphabet{\mathsfit}{bold}{\encodingdefault}{\sfdefault}{bx}{n}

\title{Large language model-powered evolutionary code optimization on a phylogenetic tree}

\author{ \href{https://orcid.org/0009-0003-3441-7381}{\includegraphics[scale=0.06]{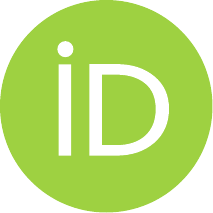}\hspace{1mm}Leyi~Zhao}\thanks{Equal Contribution.} \\
	Department of Computer Science\\
	Luddy School of Informatics\\
        Indiana University, Bloomington\\
	Bloomington, IN 47408 \\
	\texttt{leyizhao@iu.edu} \\
	\And
	\href{https://orcid.org/0000-0000-0000-0000}{\includegraphics[scale=0.06]{orcid.pdf}\hspace{1mm}Weijie~Huang$^*$} \\
	Department of Computer Science\\
	Luddy School of Informatics\\
	Indiana University, Bloomington\\
	Bloomington, IN 47408 \\
	\texttt{wh25@iu.edu} \\
        \And
	\href{https://orcid.org/0000-0000-0000-0000}{\includegraphics[scale=0.06]{orcid.pdf}\hspace{1mm}Yitong~Guo} \\
	Department of Computer Science\\
	Luddy School of Informatics\\
	Indiana University, Bloomington\\
	Bloomington, IN 47408 \\
	\texttt{yitoguo@iu.edu} \\
        \And
	\href{https://orcid.org/0000-0000-0000-0000}{\includegraphics[scale=0.06]{orcid.pdf}\hspace{1mm}Jiang~Bian} \\
	Department of Biostatistics \& Health Data Science\\
	School of Medicine\\
	Indiana University, Indianapolis\\
	Indianapolis, IN 47408 \\
	\texttt{bianji@iu.edu} \\\
        \And
	\href{https://orcid.org/0000-0000-0000-0000}{\includegraphics[scale=0.06]{orcid.pdf}\hspace{1mm}Chenghong~Wang} \\
	Department of Computer Science\\
	Luddy School of Informatics\\
	Indiana University, Bloomington\\
	Bloomington, IN 47408 \\
	\texttt{cw166@iu.edu} \\
        \And
	\href{https://orcid.org/0000-0001-7563-9915}{\includegraphics[scale=0.06]{orcid.pdf}\hspace{1mm}Xuhong~Zhang}\thanks{Corresponding Author.} \\
	Department of Computer Science\\
	Luddy School of Informatics\\
	Indiana University, Bloomington\\
	Bloomington, IN 47408 \\
	\texttt{zhangxuh@iu.edu} \\
}

\renewcommand{\shorttitle}{Large Language Model-Powered Evolutionary Code Optimization on a Phylogenetic Tree}

\hypersetup{
pdftitle={A template for the arxiv style},
pdfsubject={q-bio.NC, q-bio.QM},
pdfauthor={David S.~Hippocampus, Elias D.~Striatum},
pdfkeywords={First keyword, Second keyword, More},
}

\begin{document}
\maketitle

\begin{abstract}
	Optimizing scientific computing algorithms for modern GPUs remains a labor-intensive and iterative process, requiring repeated cycles of code modification, benchmarking, and tuning across complex hardware and software stacks. While recent work has explored large language model (LLM)–assisted evolutionary algorithms for automated code optimization, these approaches largely rely on outcome-based selection and random mutation, underutilizing the rich trajectory information generated during iterative optimization. In this work, we propose \textbf{PhyloEvolve}, an LLM-agent system that reframes GPU-oriented algorithm optimization as an \textbf{In-Context Reinforcement Learning (ICRL)} problem, enabling trajectory-conditioned, learning-based reuse of optimization experience without model retraining.\\

PhyloEvolve integrates two complementary ICRL primitives—\textbf{Algorithm Distillation} and \textbf{prompt-based Decision Transformers}—into an iterative design workflow, treating sequences of algorithm modifications and performance feedback as first-class learning signals. To structure and preserve optimization history, we introduce a \textbf{phylogenetic tree representation} that captures divergence, inheritance, and recombination among algorithm variants, supporting backtracking, cross-lineage transfer, and reproducibility. The system further combines elite trajectory pooling, multi-island parallel exploration, and containerized execution to balance exploration and exploitation across heterogeneous hardware backends.\\

We evaluate PhyloEvolve on multiple scientific computing workloads, including numerical PDE solvers, manifold learning, and spectral graph algorithms, demonstrating consistent improvements in runtime performance, memory efficiency, and correctness over baseline implementations and evolutionary search variants. Our results show that trajectory-conditioned ICRL provides a principled and effective alternative to mutation-centric evolutionary optimization, enabling scalable, interpretable, and transferable GPU code optimization for scientific computing. Our code is published at: \url{https://github.com/annihi1ation/phylo_evolve}

\end{abstract}

\keywords{In-Context Reinforcement Learning (ICRL)\and Large Language Models (LLMs) \and Agents \and Trajectory-Conditioned Learning \and Phylogenetic Tree}
\section{Introduction}

GPU-accelerated optimization has become standard in modern algorithm design. However, in scientific computing, a vast array of core algorithms (numerical Partial Differential Equations or PDEs, sparse/graph and linear algebra, combinatorial optimization, molecular/materials simulation, etc.) remain insufficiently optimized for GPUs: while the ideas are mature, GPU-side execution has not been systematically exploited, lacking cross-task and cross-backend reusable engineering loops. Inherently, effective GPU optimization is iterative: hardware characteristics are complex (memory hierarchy, warp scheduling, register allocation), the algorithm-hardware mapping space is vast, performance feedback is dense but non-analytical, thus, engineers must repeatedly ``code→benchmark→tune'' to approach hardware peak performance. Existing practices often lean heavily on experts conducting extensive manual experimentation across kernel partitioning, memory layout, parallelism granularity, and asynchronous overlapping, resulting in high development barriers and lengthy cycles. Additionally, practitioners must repeatedly balance trade-offs across different hardware and compilation/runtime environments (e.g., thread block/warp scheduling, shared memory and register allocation, tensorization instructions, and pipeline parallelism), introducing extra burdens for reproduction and portability.

This iterative optimization process naturally generates \textbf{rich optimization trajectories}: the context of each modification (historical attempts, success/failure patterns, performance bottleneck identification) constitutes \textbf{reusable optimization knowledge}—which modifications work, why they work, and under what constraints they can transfer. If this knowledge is explicitly modeled and reused, it can significantly reduce redundant exploration and accelerate convergence to high-quality solutions. Recent work has explored various approaches to leverage such optimization knowledge, including search-based methods, learned cost models, and automated tuning frameworks. Among these, LLM-assisted algorithm design has emerged as a promising direction for its ability to understand and generate code while reasoning about optimization patterns. However, existing ``LLM-assisted algorithm design/automated optimization'' approaches predominantly use Evolutionary Algorithms (\textbf{EA}) \cite{holland1975adaptation,hansen2001cmaes,deb2002nsga2,salimans2017evolution} as their backbone (e.g., LLAMEA ~\cite{vanstein2025llamea} and AlphaEvolve ~\cite{novikov2025alphaevolve}). The benefits of such adoption with EA include: these are conducive to parallelization and diversity, generating diverse algorithm variants for a given task with verifiable metrics, executing selection, mutation, and crossover based on evaluation results as fitness, forming a ``generation—evaluation—elitism—evolution'' loop. However, \textbf{trajectory-conditioned learning mechanisms around GPU execution efficiency and deployability remain weak}, relying more on operator-level random mutations rather than structured historical reuse. This dependency constrains convergence, reduces robustness, and impedes reproducibility and cross-platform transfer.

In practice, GPU iterative optimization naturally forms a ``small-step modification→immediate evaluation→history-based continuation'' loop: each modification is equivalent to the next action in a trajectory; evaluation signals for execution efficiency/correctness are dense and approximately unbiased; historical trajectories can serve as context for guidance. This setup is isomorphic to \textbf{In-Context Reinforcement Learning (ICRL)} \cite{duan2016rl2,chen2021decision,wang2016learning,lee2023supervised_icrl}. From an optimization paradigm perspective, ICRL and EA share a common structure: both center on iterative trial-and-error, moving from suboptimal to better solutions in a ``propose candidate—evaluate—improve'' loop; both can guide search with holistic performance signals without requiring explicit gradients. However, compared to EA, ICRL's learning signals are richer, sample efficiency is higher, information transfer is more continuous, and it can better leverage the model's pattern recognition and sequence understanding capabilities—capturing the contrast and complementarity between ``learning patterns from complete experience of \textbf{full trajectories}'' and ``screening by outcome scoring'',  which are critical for distinguishing ICRL from EA and for enabling more effective learning from optimization trajectories.

Based on these observations, we choose the ICRL route as our proposed GPU optimization backbone strategy, explicitly casting the ``design—implement—evaluate—redesign'' loop as a conditioned, interactive process, in which trajectories are first-class signals for policy improvement. Specifically, we frame two representative ICRL paradigms—Algorithm Distillation (\textbf{AD}) \cite{laskin2023algorithm,chen2021decision} and prompt-Decision Transformer (\textbf{prompt-DT}) \cite{xu2022promptdt,yang2024pretrained_promptdt,wang2024hierarchical_promptdt}—as our two core primitives: ``distilling algorithms from experience'' and ``guiding search with conditioned trajectories'', and embed these primitives into an LLM-driven algorithm iterative optimization workflow. We also adopt a ``phylogenetic tree'' abstraction from evolutionary biology to serve as a persistent carrier of optimization history, structuring divergence and inheritance among candidate solutions, recording lineages and performance evidence, and enabling backtracking and recombination along branches. This representation not only depicts the expansion paths of the design space but also provides a basis for subsequent transfer across node data records (code, metrics, and modification history), operator combinations, and hardware constraints.

In this paper, we propose \textbf{PhyloEvolve}: an LLM agent system based on ICRL primitives, targeting GPU-accelerated iterative design for scientific computing, enabling agents to efficiently advance algorithm design and implementation in an ICRL-like manner through a ``generation—evaluation—distillation—conditioning'' loop. The system comprises four coordinated components: (i) phylogenetic trees to carry, branch, and recombine optimization history; (ii) an Elite Pool for candidate retrieval and behavior cloning; (iii) multi-island parallel exploration to improve diversity and robustness; and (iv) containerized execution for end-to-end reproducibility and rapid backend switching (CUDA/HIP/multiple toolchains). Together, system components treat trajectories as conditioned, reusable first-class citizens, replacing operator-level random mutations with learning-based in-context transfer/alignment. The \textbf{contributions} of this work are as follows:

\begin{enumerate}
    \item \textbf{PhyloEvolve}: an ICRL-based LLM agent system using phylogenetic trees, Elite Pool, multi-island parallelism, and containerized execution for GPU-accelerated iterative design for scientific computing;
    \item \textbf{Unified ICRL adaptation}: We adapt Algorithm Distillation and prompt-Decision Transformer as ``experience distillation'' and ``conditioned trajectory search'' primitives, demonstrating ICRL as an effective alternative to EA for overcoming the limitations of manual tuning, random mutation, and poor cross-task generalization;
    \item \textbf{Empirical validation}: Across three scientific computing tasks, we validate ICRL-driven efficiency gains with PhyloEvolve.
\end{enumerate}

\section{Background and Problem Formulation}

%\todo{Rewrite}

%\subsection{LLM-Only Code Optimization}

%Approaches using only LLMs for code improvement include self-refinement prompting and RLHF-guided generation. These methods can incrementally improve code correctness or style but often lack breadth in exploration and rarely handle long-horizon optimization tasks effectively.

%\subsection{Evolutionary Computation for Code}

%Classical GP and related methods have been used in automatic programming and performance tuning, especially in compiler optimizations or low-level numerical kernels. Their common limitation lies in generic syntactic mutation strategies that frequently yield invalid or useless code fragments.

%\subsection{LLM–Evolutionary Hybrids}

%FunSearch brought LLMs into the evolutionary loop, enabling discovery of new algorithmic solutions; AlphaEvolve expanded this to large-scale, multi-objective, multi-language improvements \cite{romera2024funsearch,novikov2025alphaevolve}. Additionally, open-source frameworks like OpenEvolve and OpenELM offer extensible platforms for experimenting with LLM–EC combinations, incorporating quality-diversity search and multi-goal optimization (see e.g.\ [other works]). Our work builds on these foundations, addressing efficiency and adaptability in evolutionary prompting.

\subsection{Iterative Optimization as Trajectory Learning}
Many scientific computing and systems optimization problems—particularly GPU kernel and algorithm optimization—are inherently iterative. Rather than being solved in a single design step, practitioners repeatedly apply small modifications to an existing implementation, measure performance, and decide how to proceed based on observed outcomes. Each iteration typically involves a localized code change (e.g., modifying memory layout, kernel fusion, parallel granularity, or scheduling strategy), followed by immediate evaluation through execution, profiling, or benchmarking.

Crucially, this process generates dense, step-wise feedback: each modification yields a measurable change in runtime, memory usage, numerical stability, or correctness. Over time, these steps form optimization trajectories—ordered sequences of algorithm states, modifications, and performance deltas—that encode not only what works, but how improvements are achieved. These trajectories capture rich contextual information, including historical successes and failures, interaction effects among modifications, and constraints imposed by hardware and runtime environments.

From a learning perspective, this setting naturally aligns with sequential decision-making. Each optimization step can be viewed as an action taken in the context of the current algorithm state and its optimization history, with immediate feedback serving as a reward signal. Unlike classical reinforcement learning environments with sparse or delayed rewards, iterative algorithm optimization provides high-frequency, approximately unbiased feedback, making it particularly amenable to learning from trajectories rather than isolated outcomes.

Despite this structure, most automated optimization approaches treat iterations as independent trials or rely on population-level selection, failing to fully exploit the informational content embedded in optimization histories. This observation motivates framing iterative optimization not merely as search, but as a trajectory learning problem.

\subsection{In-Context Reinforcement Learning versus Evolutionary Algorithms} 
Evolutionary Algorithms have been a natural choice for automated algorithm and code optimization due to their simplicity, parallelism, and robustness. In EA-based approaches, candidate solutions are generated via mutation or recombination, evaluated using a fitness function, and selected based on performance. This paradigm has proven effective for exploring large, non-differentiable search spaces. However, EA fundamentally operates on outcome-based selection. Information about how a solution was obtained—i.e., the sequence of modifications leading to it—is typically discarded or weakly utilized. Mutations are often applied without conditioning on historical context, and learning occurs implicitly through population statistics rather than explicitly through trajectories. As a result, EA can suffer from inefficiencies such as redundant exploration, slow convergence, limited transferability across tasks or hardware, and weak reproducibility.

In contrast, ICRL treats trajectories as first-class learning signals. Rather than updating model parameters through gradient-based learning, ICRL induces adaptive behavior by conditioning a sequence model on past interactions—states, actions, and rewards—within the input context. The model learns to infer an implicit update rule from experience and apply it immediately to new situations, without retraining.

From a structural standpoint, EA and ICRL share a common iterative loop—propose → evaluate → improve. The key distinction lies in where learning occurs: EA learns at the population level through selection pressure while ICRL learns at the trajectory level through contextual conditioning. This difference has important implications for algorithm optimization. ICRL enables continuous information transfer across steps, allows fine-grained reuse of historical patterns, and leverages the sequence modeling capabilities of LLMs to recognize and generalize optimization strategies. As a result, ICRL can achieve higher sample efficiency and better cross-task generalization than mutation-centric evolutionary search, particularly in domains with dense feedback and strong temporal dependencies.

\subsection{Algorithm Distillation and Prompt-Based Decision Making}
Two complementary paradigms underpin ICRL-based optimization: Algorithm Distillation and prompt-based Decision Transformers. While originally developed in reinforcement learning contexts, both provide conceptual tools that generalize naturally to iterative algorithm design.

Algorithm Distillation focuses on distilling the learning process itself into a sequence model. Rather than training a model to execute a fixed policy, AD trains models to internalize update rules by observing trajectories of interaction—how actions evolve in response to rewards over time. In this view, the model does not merely learn optimal actions, but learns how to improve. This abstraction extends beyond control policies to any iterative refinement process, including algorithm and code optimization, where update heuristics play a central role.

Decision Transformers frame sequential decision-making as autoregressive sequence modeling, conditioning future actions on desired outcomes and past trajectories. Prompt-based variants extend this idea by allowing a small number of high-quality trajectories or modification sequences to serve as conditioning prompts, enabling few-shot adaptation to new tasks or objectives. Importantly, these prompts encode structured experience rather than static instructions, making them well-suited for guiding search in complex optimization landscapes.

When applied to algorithm optimization, these paradigms shift the focus from discovering isolated high-performing solutions to learning improvement strategies from experience. By conditioning on historical modification sequences, performance deltas, and target objectives, a frozen sequence model can act as an in-context policy that proposes informed, context-aware refinements. This approach enables zero-shot adaptation to new algorithms or hardware backends, relying solely on contextual demonstrations rather than retraining. Together, AD and prompt-based decision making provide the conceptual foundation for treating iterative algorithm optimization as an ICRL problem—where trajectories, rather than populations or gradients, drive learning and generalization.
\section{Related Work}
\subsection{Genetic Programming and Evolutionary Computation}
Automated algorithm design can be traced back to Genetic Programming (GP), which searches for executable solutions in program space through evolutionary operators, establishing the core paradigm of ``generate—evaluate—select'' \cite{koza1992genetic,sobania2022comprehensive}. In model selection and hyperparameter tuning, Hyperparameter Optimization (HPO) has formed a framework centered on black-box and Bayesian optimization, supplemented by multi-fidelity strategies, and has become a key component of AutoML \cite{feurer2019hyperparameter}. AutoML integrates data preparation, feature engineering, HPO, and Neural Architecture Search (NAS) end-to-end, driving NAS evolution from multi-stage/single-stage to one-shot and joint ``architecture-hyperparameter'' optimization, establishing evaluation conventions on benchmarks such as CIFAR-10 and ImageNet \cite{he2021automl}. Recent Meta-Black-Box-Optimization (MetaBBO) characterizes algorithm selection/configuration, operator design, and algorithm generation from a unified perspective, incorporating reinforcement learning, supervised learning, neuroevolution, and LLMs into a single practical framework, emphasizing the transition from experience-driven to data-driven design processes \cite{ma2024toward}.

The combination of LLMs and Evolutionary Computation (EA) has matured rapidly, with the core idea of using language models to generate candidate solutions in automated evaluation loops while employing evolutionary mechanisms to maintain diversity and elite retention. Representative advances include: In program/algorithm search, FunSearch employs a ``generate—offline evaluate—filter—regenerate'' loop to progressively explore function and program spaces, demonstrating cross-task transferability \cite{romera2023funsearch}; AlphaEvolve introduces island-based parallelism and multi-evaluator feedback, organizing robust search across tasks such as mathematics, scheduling, and system optimization through universal coding agents \cite{novikov2025alphaevolve}. 

For heuristics and metaheuristics, works such as AEL \cite{liu2023algorithm}, EoH \cite{liu2024eoh}, LLaMEA \cite{vanstein2025llamea}, and ReEvo \cite{ye2024reevo} achieve strong competitiveness on benchmarks including TSP, bin packing, and BBOB through language-generated operators, co-evolution of ``idea—code,'' systematic metaheuristic search, and reflective evolution respectively; some research directly employs LLMs as evolutionary optimizers, achieving performance comparable to classical algorithms in combinatorial optimization \cite{liu2024lmea}. At the level of evolutionary operations and prompt optimization, Language Model Crossover uses LLMs as semantic-level crossover operators for structured recombination \cite{meyerson2023language}, while EvoPrompt surpasses manual and automatic prompt baselines across multiple tasks through evolutionary search of natural language prompts \cite{guo2024evoprompt}; supporting ecosystems like OpenELM provide reproducible foundations for mutation/crossover and Quality-Diversity (QD) algorithms \cite{bradley2024openelm}. Overall, this lineage shares several mechanistic commonalities: stabilizing search through automated evaluation and elite retention, maintaining coverage through semantic mutation and QD, controlling contamination through sandboxing and unit testing, and improving sample efficiency and interpretability through reflection/self-evaluation.
\subsection{In-Context Reinforcement Learning}
In-Context Reinforcement Learning focuses on inducing policies and ``update rules'' instantly through demonstrations and feedback in long contexts without updating parameters. The AD line distills learning processes into sequence models, initially demonstrating history-based in-context learning and exploration capabilities on offline/interactive trajectories \cite{laskin2023algorithm}, with subsequent work jointly distilling dynamics/planning signals into models and extending to longer sequences and large-scale continuous control scenarios through structures like Mamba/SSM \cite{son2025dicp}. In parallel, the Prompt-DT series is based on return-conditioned modeling, enabling few-shot transfer using a small number of high-return trajectories as prompts, with improved robustness through parameter-efficient and structured prompting \cite{xu2022promptdt,hu2023ptdt,zheng2024dpdt}; theoretically, Transformers as decision makers and related work provide provable conditions for sequence models to achieve ICRL. Evidence for general LLMs is also gradually improving: in-context bandit settings \cite{monea2024banditICRL} and ``Reward Is Enough: LLMs Are In-Context RL'' \cite{song2025rewardisenough} show that iterative improvement across multiple prompting rounds is possible using only scalar (including self-generated) rewards, and ICRL has been applied to composite tasks coupling retrieval and generation (such as Text-to-SQL) for dynamic adaptation of examples and policies \cite{toteja-etal-2025-context}.

Existing LLM$\times$EA series have accumulated solid evidence in combinatorial optimization and program/heuristic discovery, but systematic methods and reproducible benchmarks for \emph{GPU kernel/operator-level} acceleration in scientific computing algorithms are still lacking. ICRL provides a unified paradigm for online adaptation without weight updates, relying solely on contextual demonstrations/preferences and runtime feedback; by viewing the iterative process of compile—run—measure as high-density, comparable reward trajectories, policy evaluation, replay, and rewriting can be completed on historical episodes. Therefore, we use ICRL as the core guiding signal, combined with evolutionary code rewriting and automated evaluation, to conduct GPU acceleration and algorithm design for scientific computing algorithms.

\subsection{Multi-Agent LLM Systems}

Recent work on multi-agent LLM systems explores how specialized agents can collaborate through role decomposition, message passing, and iterative refinement. Frameworks such as AutoGen \cite{wu2023autogen}, CAMEL \cite{li2023camel}, MATEval \cite{li2024mateval} and MetaGPT \cite{hong2023metagpt} demonstrate that structuring multiple agents with distinct responsibilities can improve planning, verification, evaluation and complex task execution. These systems primarily focus on conversational coordination or hierarchical planning in software engineering and reasoning tasks. In contrast, \tool employs a multi-agent architecture specifically to support iterative algorithm optimization: the NextStepper proposes refinement steps, the ModifyAgent verifies and executes modifications, the Designer performs structural transformations, and the Summarizer compresses trajectory history. Rather than relying on natural-language debate, agents interact through structured optimization trajectories and performance feedback, enabling coordinated exploration while maintaining correctness constraints. This design aligns with multi-agent LLM principles but is tailored to the requirements of in-context reinforcement learning and GPU-oriented scientific code optimization.
\section{Methodology}

\subsection{Overview}

\tool is an algorithm optimization framework that leverages ICRL to iteratively improve algorithms without updating model weights. Drawing inspiration from biological evolution, the system organizes algorithm variants into phylogenetic trees that capture their evolutionary history, while maintaining an elite pool of high-performing solutions to guide future exploration. Unlike traditional optimization methods that directly modify algorithms or iteratively retrain models, \tool optimizes the \emph{context} provided to frozen LLMs, enabling stable, interpretable, and transferable performance gains. 

The core innovation lies in treating algorithm optimization as an evolutionary process where the phylogenetic tree structure provides a compact representation of the algorithm's development history, with \emph{nodes} representing concrete algorithm variants (code plus metrics), \emph{edges} encoding single-step modifications and their rewards, and the \emph{forest} capturing a hierarchy of multiple trees corresponding to different algorithm families or starting designs. Elite exemplars drawn from the elite pool serve as successful demonstrations for in-context learning, and multiple specialized agents collaborate to balance exploration and exploitation. The system continuously refines its ``learning environment'' by accumulating and organizing successful modification patterns, essentially learning ``how to improve'' rather than just ``what to improve.''

\subsection{Theoretical Foundations: In-Context Reinforcement Learning}

\subsubsection{Algorithm Distillation}
\tool builds upon algorithm distillation or AD, which distills the learning process of RL algorithms into neural networks. Rather than learning a single optimal policy, our system learns to improve algorithms based on historical trajectories. We adapt AD's core principles by using learning traces as context, where historical modification sequences $(s_t, a_t, r_t)$ form the context. Here, states $s_t$ represent algorithm versions, actions $a_t$ are modifications, and rewards $r_t$ measure performance gains. The model employs sequence prediction to forecast the next beneficial modification given the evolutionary history, effectively internalizing an improvement strategy—learning the meta-strategy of optimization itself. This approach enables zero-shot adaptation, allowing new optimization tasks to be addressed by providing relevant context without retraining.

In our framework, the distillation process operates by constructing rich contextual prompts from the phylogenetic tree structure rather than through explicit weight updates. Specifically, we collect successful modification trajectories from the evolutionary history—sequences of state-action-reward tuples $([s_0, a_0, r_0], [s_1, a_1, r_1], \ldots)$ that demonstrate how algorithms progressively improve. These trajectories are formatted as structured context and provided to the frozen LLM, which then learns to predict beneficial next actions by recognizing patterns in the demonstration sequences. The key insight is that by exposing the model to multiple successful evolutionary paths, we implicitly distill the ``learning algorithm'' itself: the model internalizes not just what good algorithms look like, but how they are iteratively refined. This in-context distillation enables the LLM to generalize improvement strategies to novel algorithm states without any gradient-based training.

\subsubsection{Prompt-based Decision Making}
Following the Decision Transformer paradigm \citep{chen2021decision}, we frame algorithm optimization as autoregressive sequence modeling, where the next modification $a_t$ is predicted by:
\begin{equation}
a_t = f_\theta(R_{target}, s_t, \tau_{<t})
\end{equation}
where $a_t$ is the next modification, $R_{target}$ represents desired performance characteristics, $s_t$ is the current algorithm state, and $\tau_{<t}$ is the historical trajectory. Crucially, the combination of these three elements—$R_{target}$, $s_t$, and $\tau_{<t}$—together constitutes the \emph{prompt} provided to the LLM. This prompt encapsulates the target objective, the current algorithmic context, and the evolutionary history, enabling the model to make informed decisions about the next modification.

Unlike standard Decision Transformers that condition on the cumulative future reward— return-to-go, we use delta rewards ($\Delta r$) to capture incremental improvements, making single-step modifications more interpretable. While we depart from the return-to-go conditioning, we retain the core architectural principle of Decision Transformers: framing sequential decision-making as autoregressive sequence prediction over [target, state, history] tuples. This sequence modeling approach, combined with AD's history-based learning, constitutes our ICRL framework.

\subsection{System Architecture}

\subsubsection{Hierarchical Organization: Forest and Trees}

The system maintains a \textit{forest} structure that implements a ``genetic-algorithm-style island model'', where each phylogenetic tree represents a semi-isolated optimization lineage, and the forest as a whole enables parallel exploration with occasional cross-lineage knowledge transfer. This \textit{forest} contains multiple phylogenetic trees representing different algorithm families—distinct algorithmic approaches such as different clustering paradigms or sorting strategies—managing population diversity through parallel evolution, lightweight knowledge transfer, and adaptive scheduling. The tree structure captures each algorithm's evolutionary lineage from its root (initial version of the algorithm), while the nodes store different algorithm versions with their code, performance metrics, modification history, and constraint satisfaction status. This hierarchical structure enables the following representation:
\begin{equation}
\text{Forest} = \{T_1, T_2, ..., T_n\}, \quad T_i = (V_i, E_i, r_i, m_i)
\end{equation}
where a forest consists of multiple trees $T_1, \ldots, T_n$, representing different algorithm families. For each tree $T_i$, $V_i$ is the node set containing algorithm variants, $E_i$ is the edge set representing parent-child modification relationships, $r_i$ is the reward function mapping nodes to their performance scores, and $m_i$ is the metadata including modification descriptions and constraint satisfaction status.

\subsubsection{Phylogenetic Tree Representation}
\begin{figure}[h]
\centering
\includegraphics[scale=.6]{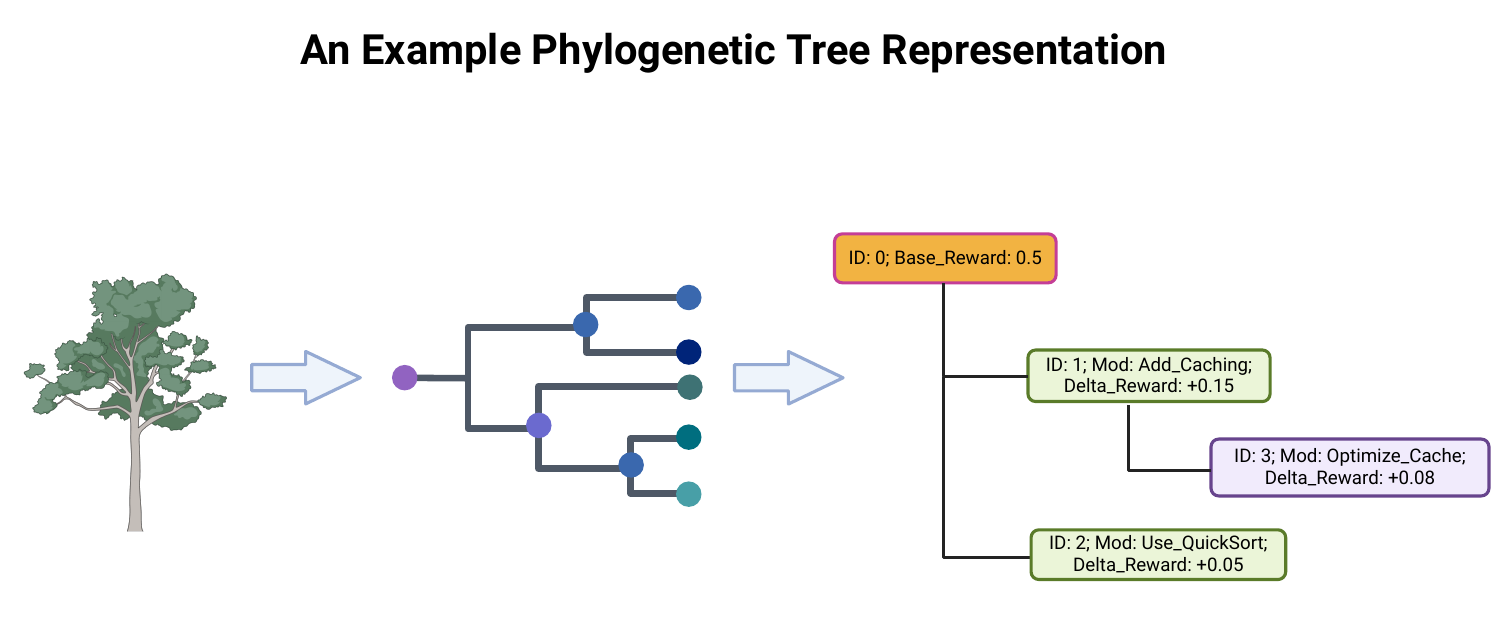}
\caption{Example of a phylogenetic tree used in \tool. Each node represents an algorithm variant, and edges correspond to refinement steps applied during optimization. The tree structure captures divergent exploration paths, where different branches pursue distinct optimization trajectories. High-performing variants contribute their trajectories and modifications to the elite pool, enabling cross-lineage reuse in subsequent iterations. This structured representation preserves diversity, prevents premature convergence, and organizes optimization history into reusable, hierarchical lineages.}
\label{fig:1}
\end{figure}

Each phylogenetic tree compactly encodes evolutionary history using \textit{S}-expressions for direct LLM consumption, where each \texttt{id} denotes a unique node (i.e., a concrete algorithm version) in the tree (as shown in Figure ~\ref{fig:1}):
%\begin{lstlisting}[language={}]
%(id:0; base_reward:0.5
%  (id:1; mod:add_caching; delta_reward:+0.15
%    (id:3; mod:optimize_cache; delta_reward:+0.08))
%  (id:2; mod:use_quicksort; delta_reward:+0.05))
%\end{lstlisting}

In this representation, the LLM can follow the complete evolutionary paths from the root node to any descendant so that it can reason about how sequences of modifications jointly affect performance, identify productive modification chains that repeatedly yield positive delta rewards, learn from abandoned branches and characteristic failure patterns (e.g., sequences that systematically degrade reward or violate constraints), and ultimately generalize these learned strategies to new but structurally similar contexts by matching the current partial trajectory to previously observed paths.

\subsubsection{Elite Pool: Cross-Lineage Knowledge Transfer}

The Elite Pool serves as a global memory mechanism that enables explicit cross-lineage knowledge transfer across the phylogenetic forest. While individual phylogenetic trees evolve largely independently to preserve diversity, the elite pool operates across all trees, allowing successful optimization experience discovered in one lineage to inform decision-making in others. Specifically, the elite pool maintains two complementary forms of reusable optimization knowledge:
\begin{itemize}
    \item \textbf{Elite trajectories}. First, the pool stores elite trajectories, which are complete or partial evolutionary paths corresponding to high-performing algorithm variants. Each trajectory encodes an ordered sequence of algorithm states, applied modifications, and associated delta rewards, capturing how performance improvements were achieved over time rather than only their final outcomes. Trajectories are admitted into the pool based on top-$k$ performance criteria across the entire forest, independent of their originating tree.

    When constructing context for the \textit{NextStepper} agent (see Section ~\ref{sec:MAC}), the system retrieves a subset of elite trajectories using similarity-based sampling. Importantly, this retrieval ignores tree boundaries: trajectories originating from different phylogenetic trees may be selected and provided as demonstrations. This mechanism enables cross-lineage transfer by allowing optimization strategies learned in one algorithm family or design lineage to guide search decisions in another, even when their evolutionary histories are otherwise disjoint.    
    \item \textbf{Elite modifications.} Second, the pool maintains a catalog of elite modifications, representing individual code edits or optimization actions that have repeatedly yielded positive performance gains across different contexts. For each modification $m$, the system aggregates statistics across all trees, including its mean performance gain $\mu(m)$, variance $\sigma_{std}(m)$, and application count $n(m)$. These statistics are used to compute a value score that balances effectiveness, consistency, and confidence:    
    \begin{equation}
    \text{value}(m) = \mu(m) \cdot \left(0.5 + 0.5\sigma\left(-var(m) + \log(1+n(m))\right)\right)
    \end{equation}
    where $\sigma(\cdot)$ denotes the sigmoid function. Modifications with high value scores are prioritized during context construction, providing reliable, low-level heuristics that generalize across lineages.
\end{itemize}
Together, elite trajectories and elite modifications allow the system to transfer optimization knowledge at multiple levels of granularity: full improvement strategies through trajectory reuse, and localized, statistically validated heuristics through modification reuse. This transfer occurs in-context, without parameter updates, and is mediated entirely through the construction of prompts supplied to the \textit{NextStepper} agent.

Conceptually, the elite pool plays a role analogous to migration in island-model evolutionary algorithms, but operates at the level of trajectory-conditioned learning rather than population replacement. By separating lineage-specific exploration from global experience reuse, the elite pool enables PhyloEvolve to maintain diversity while accelerating convergence through systematic reuse of optimization knowledge across the forest.

\subsection{Multi-Agent Collaboration}\label{sec:MAC}

\subsubsection{Agent Specialization}

The system employs four specialized agents: \textit{NextStepper}, \textit{ModifyAgent}, \textit{Designer}, and \textit{Summarizer}. Each is designed to handle specific aspects of the evolutionary optimization process. This division of labor enables efficient context utilization and maintains clear separation of tasks and goals throughout the optimization pipeline.

\subsubsubsection{NextStepper: The In-Context Policy}

\textit{NextStepper} serves as the core decision-making agent, functioning as the system's ``in-context policy'' generator. At each iteration, it receives the current evolutionary context and proposes the next modification to explore. Unlike traditional RL policies that are learned through weight updates, \textit{NextStepper} operates purely through context understanding and generation, leveraging the LLM's pattern recognition capabilities to identify promising directions.

\textbf{Context Construction:} \textit{NextStepper}'s decision-making is grounded in a rich, structured context that encompasses multiple sources of information. The current trajectory provides the complete path from root to the focal node, encoded as a sequence of modifications and their delta rewards, offering immediate and past evolutionary history. Sibling comparisons reveal the performance and modification patterns of neighboring branches, providing near-field insights into alternative paths and their relative merits. By conducting sibling comparisons, \textit{NextStepper} can become aware of which approaches have succeeded or failed in similar contexts. The context also incorporates elite trajectories—high-performing complete paths of the exemplars from the elite pool that serve as successful candidates demonstrating efficient and better modification sequences according to the given metrics. Additionally, elite modifications provide statistically validated single-step edits with proven effectiveness, offering reusable local heuristics. Finally, task constraints including resource limits, evaluation protocols, and domain-specific requirements bound the feasible modification space. Together, these elements provide \textit{NextStepper} with both local and global perspectives on the optimization landscape, integrating historical, comparative, and exemplary knowledge to guide future steps.

\textbf{Mode-Adaptive Behavior:} \textit{NextStepper} operates in three distinct modes, each tailored to different phases of the optimization process:

\begin{itemize}
    \item \textbf{Warmup Mode.} During the initial phase, \textit{NextStepper} prioritizes safety and diversity over aggressive optimization. It proposes conservative, low-risk micro-edits such as parameter adjustments, minor algorithmic refinements, or localized optimizations. The goal is to establish a robust initial frontier while calibrating the system's understanding of task constraints and evaluation behavior. Context weighting is applied and it emphasizes the current trajectory and near-field comparisons, downweighting distant elite exemplars to avoid premature convergence to potentially domain-mismatched patterns.
    \item \textbf{Explore Mode.} When performance plateaus or diversity metrics indicate convergence, \textit{NextStepper} shifts to exploration. In this mode, it actively seeks modifications that diverge from previously explored paths, introducing alternative mechanisms, data structures, or algorithmic paradigms. The system increases the influence of diverse elite exemplars and allows larger modification step sizes while maintaining clear rollback paths. Exploration proposals explicitly articulate novelty justifications and validation strategies to ensure a controlled and principled expansion of the search space.
    \item \textbf{Exploit Mode.} When elite memory provides clear directional signals and recent modifications show consistent gains, \textit{NextStepper} enters exploitation mode. It proposes small, verifiable refinements around proven patterns, focusing on incremental performance consolidation. Interface stability is prioritized to maintain comparability across iterations, and modifications target specific bottlenecks identified in successful branches. The system emphasizes reusing high-value elite modifications and following trajectory patterns that exhibit sustained improvement.
\end{itemize}

These adaptive modes enable \textit{NextStepper} to dynamically adjust its strategy based on the current state of the optimization process, balancing between establishing a stable foundation, exploring new possibilities, and refining successful approaches. This dynamic adaptation ensures the system remains effective across different stages of the optimization lifecycle.

\textbf{Output Structure:} \textit{NextStepper} produces a two-layer output consisting of a high-level proposal and detailed specification. The high-level proposal provides a semantic summary suitable for tree logging, retrieval, and cross-lineage transfer, including the modification intent, expected performance impact, and reasoning based on context patterns (e.g., ``similar to the successful caching strategy in Tree $i$''). The detailed specification contains an execution-ready description with specific code locations, transformation rules, and validation criteria for the \textit{ModifyAgent} to implement (e.g., ``insert cache check at line $45$''). Additionally, each proposal includes a brief analysis covering expected gains, potential risks, and fallback strategies, which feeds back into future context construction. This output structure leads to the creation of a new child node in the phylogenetic tree.

\subsubsubsection{ModifyAgent: The Executor for Precision} 

\textit{ModifyAgent} bridges the gap between high-level proposals and executable code changes. It operates under strict principles of minimal invasiveness, maintaining system stability while implementing modifications. \textit{ModifyAgent} receives the detailed specification from \textit{NextStepper} as input.

\textbf{Code Generation Strategy:} \textit{ModifyAgent} follows a ``minimal diff'' philosophy that emphasizes localized changes. Modifications are scoped to the smallest possible code regions, avoiding unnecessary refactoring or style changes. The agent preserves invariants by maintaining external interfaces, API contracts, and evaluation protocols stable across modifications to ensure comparability and enable knowledge reuse across lineages. Furthermore, each edit has well-defined entry and exit points, establishing boundary clarity that make rollback straightforward and enables fine-grained performance attribution.

\textbf{Validation and Verification:} Before accepting a modification, \textit{ModifyAgent} performs comprehensive validation including syntax and type checking to catch immediate errors, constraint verification against task-specific requirements such as time/space complexity bounds, execution in a sandboxed environment with resource monitoring, and comparative evaluation against the parent node to compute delta reward.

\textbf{Debugging and Error Recovery:} When execution fails or performance regresses, \textit{ModifyAgent} enters an automated debugging workflow: First, it performs error diagnosis by parsing execution traces, error messages, and constraint violations to identify the root cause. Next, it generates targeted repairs with minimal fixes that address the specific failure mode without altering unrelated code, prioritizing preservation of the original modification intent. The agent then retries with limits, attempting execution again with the repaired code; if multiple repair attempts fail, it escalates to rollback or flags for \textit{Designer} intervention. Finally, it documents failures by explicitly marking failed paths in the tree with error summaries, creating negative examples that inform future decisions.

\subsubsubsection{Designer: The Structural Innovator}

\textit{Designer} addresses situations where incremental modifications fail to overcome fundamental limitations. It operates less frequent than \textit{NextStepper} and \textit{ModifyAgent}, intervening only when systematic issues require architectural changes.

\textbf{Redesign Strategy:} Unlike incremental modifications, \textit{Designer} performs holistic transformations through architectural refactoring, algorithmic paradigm shifts, and information flow optimization. Architectural refactoring restructures module boundaries, decomposes monolithic components, or introduces new abstraction layers to enable different optimization trajectories. Algorithmic paradigm shifts replace core algorithmic strategies, such as transitioning from dynamic programming to greedy heuristics or from iterative to recursive approaches, when current paradigms show inherent limitations. Information flow optimization reorganizes data structures, communication patterns, or control flow to eliminate bottlenecks identified through accumulated evidence.

\textbf{Context Utilization:} \textit{Designer} leverages deep evolutionary memory by starting from the highest-performing node in the elite pool as a foundation (triggered by the conditions described above). It analyzes trajectory summaries to identify preserved capabilities and common failure modes, retrieves semantically similar elite trajectories from diverse lineages to inform redesign principles, and synthesizes patterns across multiple successful branches to extract transferable design heuristics.

\textbf{Integration with Evolution:} \textit{Designer} outputs are treated as new tree roots in the forest. If a redesigned algorithm passes performance thresholds (e.g., exceeds the median reward of existing tree roots), it is admitted as a new lineage. This creates opportunities for cross-pollination: subsequent \textit{NextStepper} iterations can sample and refine the new paradigm, while elite memory preserves patterns from both old and new lineages.

\subsubsubsection{Summarizer: The Knowledge Distiller}

\textit{Summarizer} operates as a meta-learning component, periodically compressing successful trajectories into reusable insights that enhance future decision-making.

\textbf{Trajectory Analysis:} \textit{Summarizer} examines complete evolutionary paths to extract productive modification sequences—patterns of edits that consistently lead to performance gains such as ``caching before optimization'' or ``data structure selection before algorithmic refinement.'' It identifies constraint-aware strategies that successfully navigate task-specific constraints, documents failure patterns including common pitfalls and anti-patterns to avoid in future iterations, and discovers transferable principles comprising domain-agnostic heuristics that generalize across different algorithm families.

\textbf{Semantic Indexing:} Summaries are embedded into a semantic space for efficient retrieval. During \textit{NextStepper}'s context construction, the top-$k$ most relevant summaries (by cosine similarity to the current trajectory) are included, providing high-level guidance without overwhelming the context window.

\textbf{Incremental Refinement:} As the forest evolves, \textit{Summarizer} continuously updates its knowledge base through several mechanisms. New successful trajectories are analyzed and merged with existing summaries, statistical profiles including frequency, average gain, and variance are updated for each identified pattern, and low-value or redundant summaries are pruned to maintain efficiency.

\subsubsubsection{Agent Interaction Protocol}

\begin{figure}[h]
\centering
\includegraphics[scale=.4]{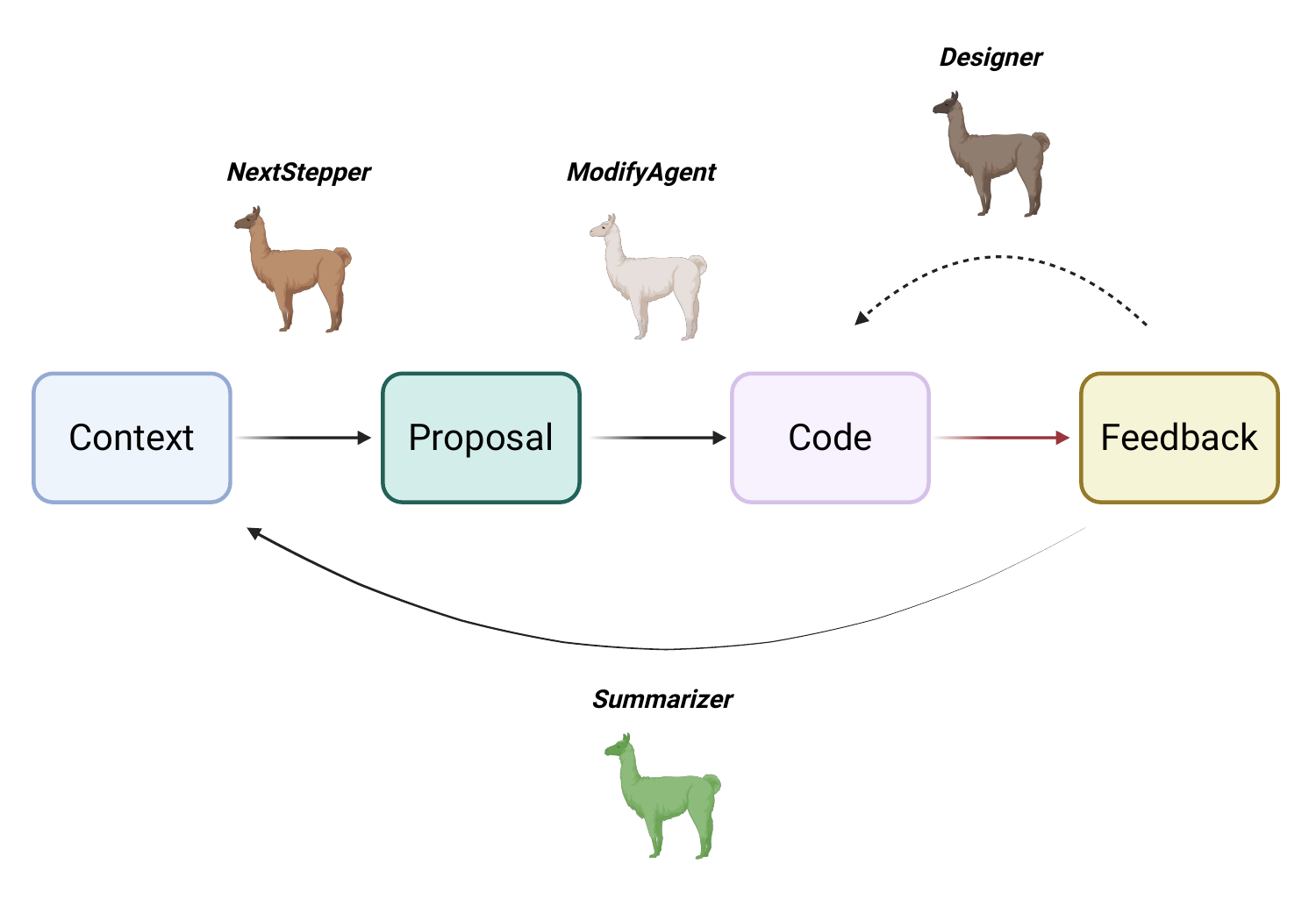}
\caption{The interaction loop between agents in \tool. The \textit{NextStepper} proposes refinement actions conditioned on historical trajectories; the \textit{ModifyAgent} applies and validates the proposed code edits; the \textit{Designer} performs higher-level structural transformations only when necessary; and the \textit{Summarizer} compresses the evolving trajectory history into reusable context. Performance feedback from executing each candidate informs subsequent proposals, and successful trajectories are routed into the elite pool. Through this coordinated multi-agent loop, the system iteratively generates, evaluates, and preserves algorithmic improvements.}
\label{fig:2}
\end{figure}

The four agents collaborate through a structured pipeline that forms a continuous optimization loop. At each iteration, \textit{NextStepper} analyzes the current evolutionary context and proposes a modification strategy; \textit{ModifyAgent} translates this proposal into executable code and validates the result; the evaluation outcome updates the phylogenetic tree; and \textit{Summarizer} periodically distills successful patterns into reusable knowledge that enriches future contexts. This flow can be summarized as in Figure~\ref{fig:2}: 
%\begin{equation}
%\text{Context} \xrightarrow{\text{NextStepper}} \text{Proposal} \xrightarrow{\text{ModifyAgent}} \text{Code} \xrightarrow{\text{Evaluation}} \text{Feedback} \xrightarrow{\text{Summarizer}} \text{Context}
%\end{equation}
%\textit{Designer} operates outside this regular loop, intervening only when the system detects structural barriers to progress (as described in the Designer section above). When triggered, \textit{Designer} creates new tree roots that join the forest, providing fresh starting points for the regular agent loop to refine.{\color{red}Delete?}

\subsection{Orchestration and Optimization}

The orchestration process coordinates the interaction between agents and the management of the phylogenetic forest. It ensures a balance between exploring new algorithm families and exploiting successful modifications within existing lineages. The main mechanisms include an adaptive sampling strategy for selecting which nodes to expand, and pruning strategies that remove low-potential candidates while preserving informative failures and maintaining diversity.

\subsubsection{Adaptive Sampling Strategy}

Not every node in the phylogenetic tree automatically grows children; instead, at each iteration, the system samples a single node to serve as the parent for the next modification attempt. This selective expansion focuses computational resources on the most promising regions of the search space. Node selection balances performance and exploration potential through:
\begin{equation}
p(n_i) \propto \exp\left(\frac{\alpha \cdot r_i + \beta \cdot \Delta r_i + \gamma \cdot d_i^{-1}}{T}\right)
\end{equation}
where $p(n_i)$ is the probability of node $n_i$ being sampled for expansion, $r_i$ is the absolute reward (overall performance), $\Delta r_i$ is the recent improvement (momentum signal), $d_i$ is the depth in the tree (shallower nodes are favored for broader exploration), and $T$ is a temperature parameter controlling exploration randomness. Higher temperatures produce more uniform sampling, while lower temperatures concentrate selection on high-scoring nodes.

Forest-level sampling determines which tree receives the next iteration, using a composite scoring function:
\begin{equation}
\text{score}(T_i) = w_1 \cdot \text{perf}(T_i) + w_2 \cdot \text{potential}(T_i) + w_3 \cdot \text{diversity}(T_i)
\end{equation}
where $\text{perf}(T_i)$ is the best reward achieved in tree $T_i$, $\text{potential}(T_i)$ measures recent improvement trends, and $\text{diversity}(T_i)$ captures how distinct $T_i$ is from other trees. This composite scoring ensures resources are allocated to trees that are either high-performing, showing promise, or contributing unique algorithmic approaches.

\subsubsection{Pruning Mechanisms}

Multi-stage pruning maintains tractability while preserving diversity, targeting nodes, branches, or entire trees to remove redundancy and failed candidates. The system employs hopeless branch removal to prune subtrees whose descendants all have $\Delta r < 0$, selective failure retention to keep the most informative failures as learning examples, and low-potential cleaning to eliminate stagnant, positive-gain nodes with minimal absolute reward.

Forest pruning is triggered when the number of trees $|F|$ exceeds a capacity threshold $\tau$. Trees are evaluated using a composite retention score:
\begin{equation}
\text{retain}(T_i) = \alpha \cdot \text{norm}(r_{best}(T_i)) + \beta \cdot \text{norm}(r_{weighted}(T_i)) + \gamma \cdot \text{potential}(T_i)
\end{equation}
where $r_{best}(T_i)$ is the highest reward achieved by any node in tree $T_i$, $r_{weighted}(T_i)$ is a depth-weighted average reward across the tree (giving more weight to recent nodes), $\text{potential}(T_i)$ measures the tree's recent improvement trajectory, and $\text{norm}(\cdot)$ normalizes values across all trees. Trees with the lowest retention scores are pruned first, removing algorithm families that are neither high-performing nor showing improvement potential.

\subsubsection{Main Orchestration Loop}
\begin{figure}[h]
\centering
\includegraphics[scale=.5]{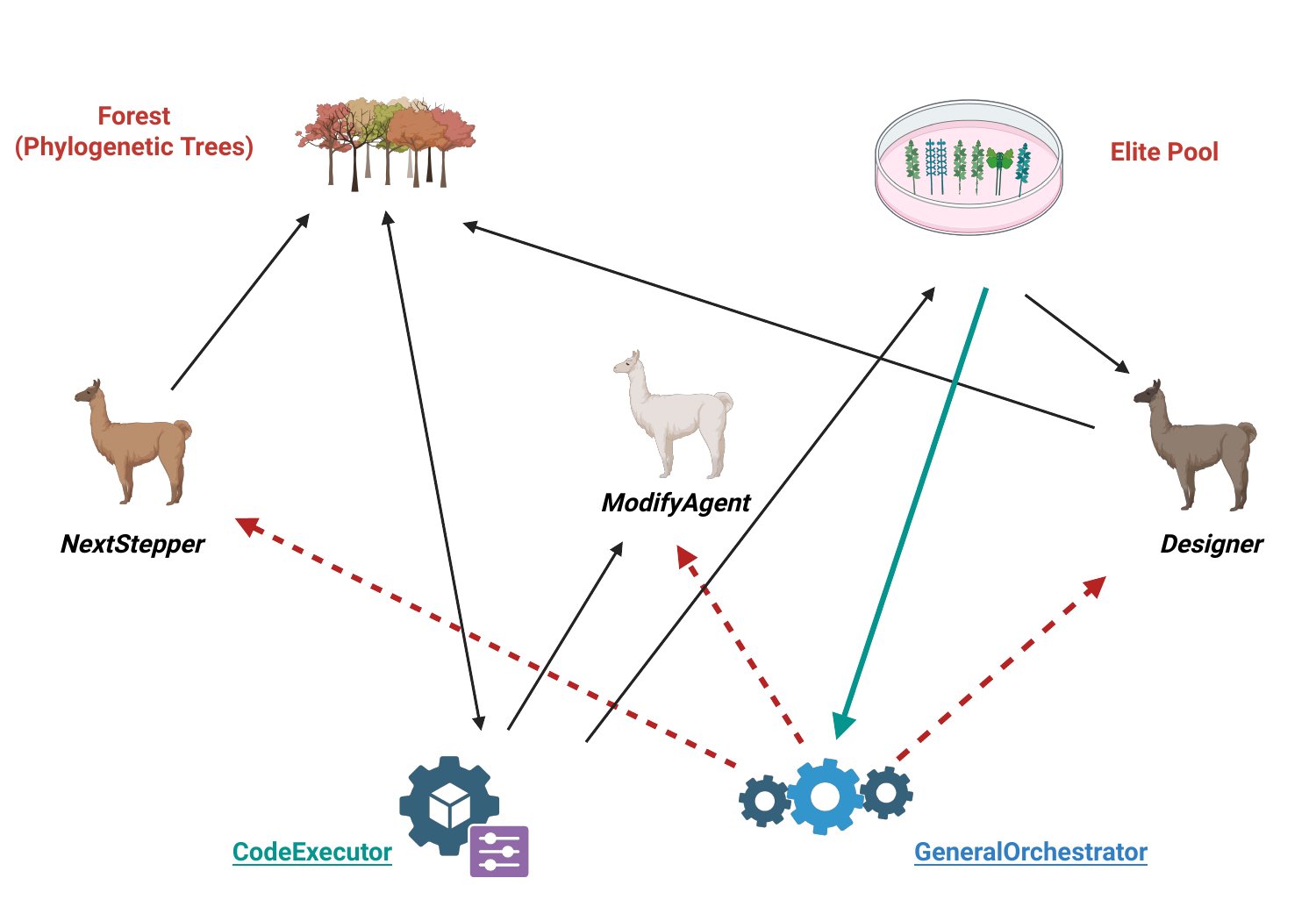}
\caption{Simplified pipeline of \tool. The Orchestrator coordinates the LLM agents to generate refinement candidates and larger structural redesigns. Generated variants are inserted into the phylogenetic forest, executed and benchmarked by the CodeExecutor, and evaluated for performance. High-performing variants are preserved in the Elite Pool, which in turn guides subsequent evolution through trajectory-conditioned retrieval and cross-lineage knowledge transfer.}
\label{fig:3}
\end{figure}

We summarize the orchestration loop in Algorithm~\ref{alg:phyloevolve}, and the whole pipeline is shown in Figure ~\ref{fig:3}.
\begin{algorithm}[h]
\caption{\tool Main Orchestration}
\label{alg:phyloevolve}
\begin{algorithmic}[1]
\State \textbf{Initialize} forest $F$ with seed algorithms
\State \textbf{Initialize} elite pool $E \leftarrow \emptyset$
\For{epoch $= 1$ to $N$}
    \State // \textit{Micro-loop: Incremental evolution}
    \State $T_i \sim$ SampleTree($F$, strategy)
    \State $n_j \sim$ SampleNode($T_i$, temperature)
    \State context $\leftarrow$ BuildContext($n_j$, $E$, $T_i$)
    \State proposal $\leftarrow$ NextStepper(context, mode)
    \State code, reward $\leftarrow$ ModifyAgent(proposal, $n_j$)
    \If{reward $> 0$}
        \State AddChild($T_i$, $n_j$, code, reward)
        \State UpdateElite($E$, trajectory, modification)
    \EndIf
    \State PruneTree($T_i$, budget)
    
    \State // \textit{Macro-loop: Structural innovation}
    \If{TriggerRedesign(history, diversity)}
        \State elite\_node $\sim$ SelectElite($E$)
        \State redesign $\leftarrow$ Designer(elite\_node, context)
        \State new\_tree $\leftarrow$ Execute(redesign)
        \If{Performance(new\_tree) $>$ threshold}
            \State AddTree($F$, new\_tree)
            \State PruneForest($F$, capacity)
        \EndIf
    \EndIf
\EndFor
\State \textbf{Return} best algorithm from $F$
\end{algorithmic}
\end{algorithm}

%\subsection{Key Innovations and Advantages}

The proposed framework introduces several key innovations that collectively redefine how optimization, memory, and exploration are approached in evolving systems. \textbf{First}, it shifts the focus from traditional weight optimization to context optimization, emphasizing adaptive coordination among modules rather than static parameter tuning. Unlike traditional approaches that update model parameters, \tool optimizes the context provided to frozen LLMs. This approach offers stability through no risk of catastrophic forgetting or training instabilities, interpretability via a complete audit trail of optimization decisions, transferability as learned patterns readily apply to new domains, and efficiency since no gradient computation or backpropagation is required. \textbf{Second}, it incorporates evolutionary memory and knowledge reuse, enabling the system to accumulate and leverage prior experience through structured retention of high-value modifications and informative failures. The phylogenetic tree structure and elite pool create a persistent evolutionary memory that preserves complete optimization trajectories for analysis, enables semantic retrieval of relevant past experiences, facilitates cross-lineage knowledge transfer, and prevents repeated exploration of failed paths. \textbf{Finally}, it achieves a balanced exploration–exploitation dynamic, where diversity is actively maintained while convergence is guided toward promising regions of the search space, ensuring both innovation and stability in long-term evolution. The multi-agent architecture with adaptive sampling naturally balances local refinement through incremental modifications for exploitation, global search through structural redesigns for exploration, diversity maintenance through forest-level population management, and risk management through rollback capabilities and constraint checking.    
%\begin{itemize}
%    \item \textbf{Context Optimization vs. Weight Optimization.} Unlike traditional approaches that update model parameters, \tool optimizes the context provided to frozen LLMs. This approach offers stability through no risk of catastrophic forgetting or training instabilities, interpretability via a complete audit trail of optimization decisions, transferability as learned patterns readily apply to new domains, and efficiency since no gradient computation or backpropagation is required.    
%    \item \textbf{Evolutionary Memory and Knowledge Reuse.} The phylogenetic tree structure and elite pool create a persistent evolutionary memory that preserves complete optimization trajectories for analysis, enables semantic retrieval of relevant past experiences, facilitates cross-lineage knowledge transfer, and prevents repeated exploration of failed paths.
%    \item \textbf{Balanced Exploration-Exploitation.} The multi-agent architecture with adaptive sampling naturally balances local refinement through incremental modifications for exploitation, global search through structural redesigns for exploration, diversity maintenance through forest-level population management, and risk management through rollback capabilities and constraint checking.    
%\end{itemize}

%\input{figs/orchastrator_v2}

\section{Experiments}

%We evaluate \tool to demonstrate its effectiveness in accelerating GPU kernel optimization and its generalization capability across hardware backends and problem domains. Our experiments are designed to answer three key questions: (1) Can ICRL-based optimization outperform traditional evolutionary or manual tuning in efficiency and performance? (2) How does the integration of trajectory-conditioned learning improve cross-search transferability and reproducibility? and (3) Can the system maintain stability and scalability across diverse computational workloads?

To comprehensively evaluate the versatility of \tool, we conduct experiments across three representative algorithmic domains: Landau–Lifshitz–Gilbert (LLG) equation solving \cite{lakshmanan2011llg}, Local Tangent Space Alignment (LTSA) \cite{zhang2004ltsa} for manifold learning, and GraphWave \cite{donnat2018graphwave} for graph spectral embedding. These benchmarks span diverse computational paradigms—numerical PDEs, geometric learning, and spectral graph analysis—covering both dense and sparse workloads. For each task, we compare \tool’s ICRL-driven optimization against baseline implementations. %and evolutionary search variants, measuring runtime efficiency, convergence behavior, and transferability across GPU backends. This setup allows us to assess not only absolute performance gains but also how effectively trajectory-conditioned learning generalizes across heterogeneous optimization landscapes.

\subsection{Setup}

All experiments were conducted on a Linux workstation equipped with an NVIDIA GPU A40, and evaluations were executed inside Docker containers using CUDA-enabled images to provide GPU acceleration with consistent, isolated environments. The framework is implemented in Python 3.11 (tooling pinned in the project configuration) and runs each candidate in a controlled sandbox with enforced timeouts and memory limits. Driven by GPT-5 for planning and design (NextStepper and Designer agents), Claude Sonnet 4 for code modification (ModifyAgent), and GPT-4.1 for summarization, the system evaluates performance via wall-clock runtime measured around each container execution and verifies correctness through a standardized run-time contract: each execution returns a structured success/failure signal with an accompanying reason on failure and a scalar score on success. The evaluation follows a two-phase protocol (warmup exploration followed by evolutionary optimization), with periodic checkpointing/pruning of the search state and up to three automatic debug retries for failing candidates.

%\subsection{Top-K Retrieval}
%Top-K retrieval is a fundamental operation in large-scale information retrieval and machine learning systems, where the goal is to identify the K most relevant items from a large candidate set according to a similarity or scoring function. It underpins a wide range of high-impact applications, including recommendation systems, search engines, nearest-neighbor retrieval, and vector database querying in modern embedding-based models. At its computational core, Top-K retrieval involves intensive vector similarity computation, partial sorting, and index scanning, which together present substantial opportunities for performance optimization on GPUs. In particular, the scoring and selection kernels—such as dot-product accumulation, heap-based partial sorting, and segment reduction—can be highly parallelized but are often memory-bound and sensitive to kernel fusion strategies. Within PhyloEvolve, we use Top-K retrieval as a benchmark for optimizing these numerical components: our framework automatically explores GPU execution configurations, refines memory layout strategies, and tunes thread-block granularity through trajectory-conditioned reinforcement learning, enabling efficient reuse of optimization knowledge across similar search and ranking workloads.

\begin{figure}[h]
\centering
\begin{subfigure}{0.48\textwidth}
\includegraphics[width=\textwidth, height=6cm]{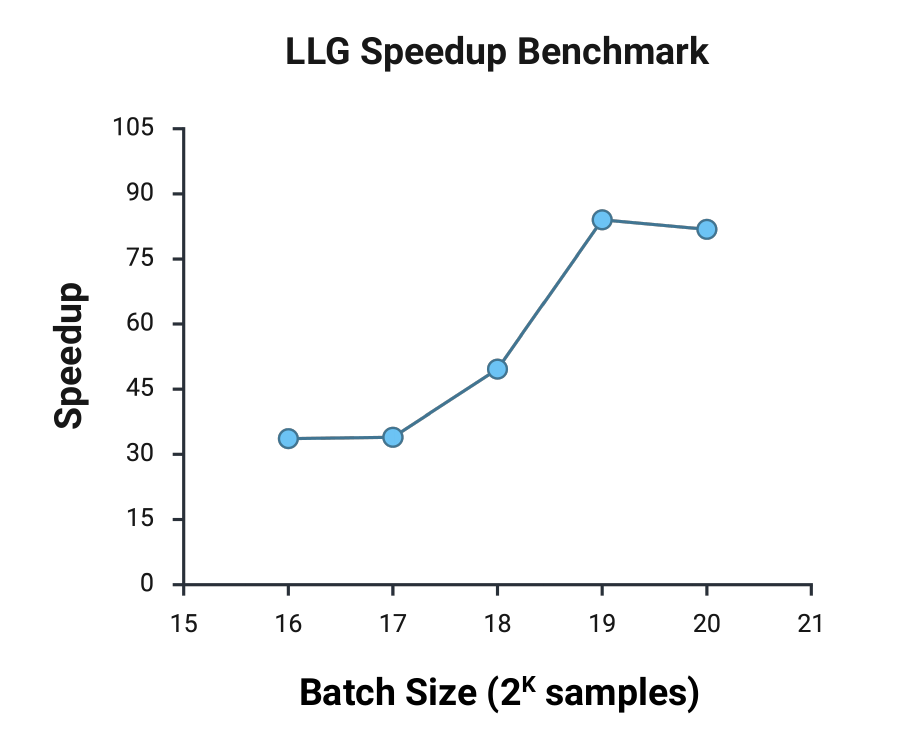} 
\caption{Speedup achieved by PhyloEvolve on the Landau–Lifshitz–Gilbert (LLG) equation solver across varying batch sizes. The optimized kernels generated by PhyloEvolve significantly outperform the baseline implementation, with speedup increasing as batch size scales. This reflects PhyloEvolve’s ability to restructure memory access patterns and reduce redundant intermediate computations in stiff, numerically intensive PDE workloads.}
\label{fig:subim1}
\end{subfigure}
\hfill
\begin{subfigure}{0.48\textwidth}
\includegraphics[width=\textwidth, height=6cm]{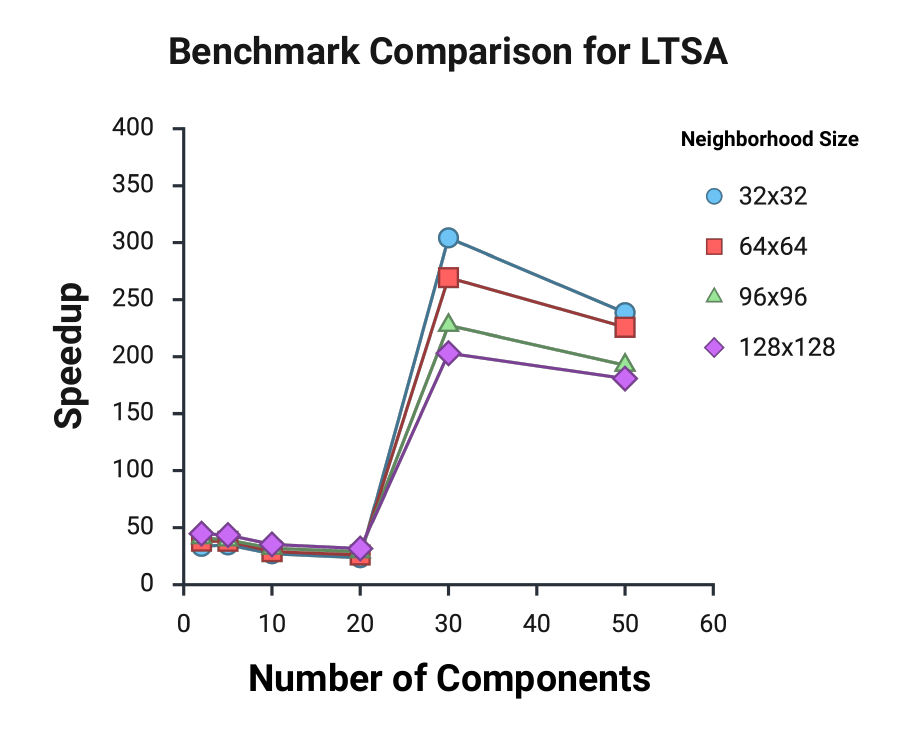}
\caption{Speedup achieved by PhyloEvolve on the Local Tangent Space Alignment (LTSA) algorithm as a function of the number of target components. PhyloEvolve consistently accelerates neighbor search, covariance estimation, and eigenvector computations, demonstrating strong performance gains across dimensionalities. The performance improvement correlates with reduced kernel launch overhead and more efficient use of GPU shared-memory for small matrix operations.}
\label{fig:subim2}
\end{subfigure}
\caption{Runtime acceleration achieved by PhyloEvolve on (a) the Landau–Lifshitz–Gilbert (LLG) PDE solver and (b) the Local Tangent Space Alignment (LTSA) manifold-learning algorithm. In both cases, PhyloEvolve generates GPU-optimized code variants that substantially outperform the baseline implementations. The system adapts to the computational structure of each workload—stiff vector-field updates for LLG and dense/small-matrix linear algebra for LTSA—achieving multi-fold speedups without compromising numerical correctness. These results highlight PhyloEvolve’s ability to discover workload-specific optimization strategies through trajectory-conditioned learning rather than random mutation.}
\label{fig:image2}
\end{figure}

\subsection{Landau–Lifshitz–Gilbert (LLG) equation}
The LLG equation governs the time evolution of magnetization in ferromagnetic materials, forming the foundation of micromagnetics and spintronics research. It models how magnetic moments precess and relax under the influence of external and internal effective magnetic fields, capturing both conservative (precessional) and dissipative (damping) dynamics. This equation is central to applications such as magnetic memory design, spin-wave simulation, and domain wall motion, making it one of the most impactful models in computational physics. Numerically, the LLG equation is a stiff, nonlinear vector PDE, often solved through iterative updates involving vector cross-products, normalization, and field coupling across spatial grids. These operations exhibit fine-grained parallelism and high memory intensity, presenting an ideal testbed for GPU optimization. Within \tool, we employ the LLG solver to benchmark how trajectory-conditioned reinforcement learning accelerates the tuning of kernel fusion, thread-block partitioning, and memory layout strategies, demonstrating how learned optimization trajectories can generalize across PDE-based workloads.

\subsection{Local Tangent Space Alignment (LTSA)}
LTSA is a nonlinear dimensionality reduction algorithm that reconstructs global manifold structure by aligning local tangent spaces estimated from high-dimensional data. It has been highly influential in manifold learning, representation discovery, and data visualization, offering a mathematically elegant way to preserve local geometry while uncovering low-dimensional embeddings. LTSA’s core computations involve neighbor search, local covariance decomposition, and global eigenvalue alignment, which are both memory-intensive and dominated by matrix–matrix operations and sparse linear algebra. These numerical components make LTSA a strong candidate for GPU optimization. Within \tool, we use LTSA to benchmark how trajectory-conditioned learning accelerates kernel optimization for neighbor-based matrix operations, eigendecomposition, and sparse matrix assembly. This task demonstrates the framework’s ability to capture and reuse optimization trajectories in high-dimensional geometric learning workloads, bridging scientific computing and machine learning applications.

%\begin{figure}[h]
%\includegraphics[scale=.5]{figs/LSTA_result.pdf}
%\end{figure}

\subsection{Spectral Wavelets for learning structural signatures in complex networks (GraphWave)}
GraphWave is a spectral graph embedding algorithm that characterizes structural roles of nodes by simulating wave propagation on graphs. It computes embeddings by applying spectral graph wavelets to the eigen-decomposition of the graph Laplacian, capturing how localized disturbances diffuse across the network. This method has been impactful in network representation learning, community detection, and graph-based pattern recognition, providing a principled spectral alternative to purely neural embeddings. From a computational perspective, GraphWave is dominated by eigendecomposition, matrix exponentiation, and spectral filtering, all of which are compute- and memory-intensive operations that map naturally to GPU acceleration. Within \tool, we employ GraphWave as a representative graph spectral workload, optimizing its spectral convolution kernels, sparse matrix–vector operations, and exponential filtering procedures. Through trajectory-conditioned reinforcement learning, the system learns to refine kernel partitioning, memory reuse, and thread scheduling across repeated spectral transforms, demonstrating its adaptability to irregular, graph-structured numerical computations.

\begin{figure}[h]

\begin{subfigure}{0.48\textwidth}
\includegraphics[width=\textwidth, height=6cm]{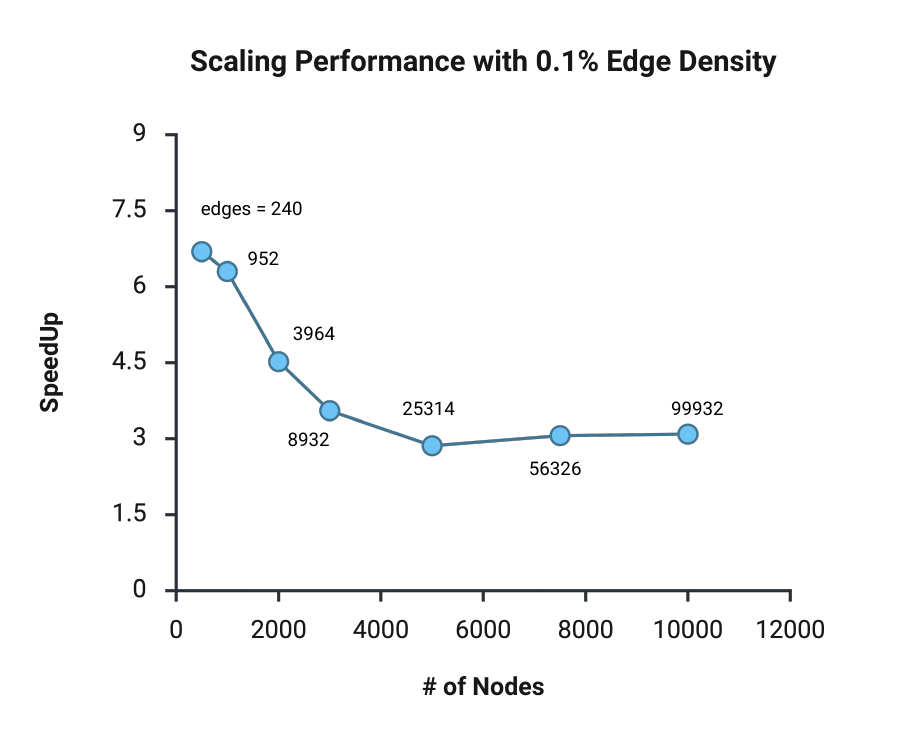} 
\caption{Speedup of PhyloEvolve-generated GPU kernels on GraphWave as a function of graph size. While small graphs exhibit the highest acceleration over the baseline, the relative speedup decreases as graph size grows, stabilizing at a consistent performance ratio for the largest graphs. This reflects the increasing dominance of eigendecomposition and spectral propagation costs at scale, where PhyloEvolve still improves GPU utilization but cannot eliminate the inherent complexity of large sparse spectral operations.}
\label{fig:subim1}
\end{subfigure}
\hfill
\begin{subfigure}{0.48\textwidth}
\includegraphics[width=\textwidth, height=6.1cm]{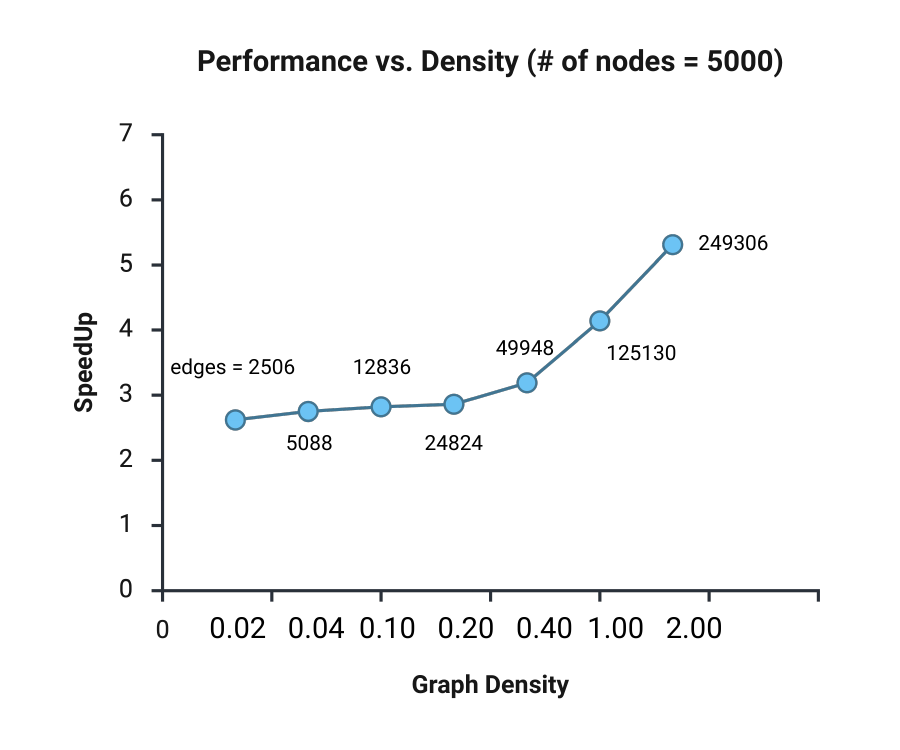}
\caption{Speedup of PhyloEvolve on GraphWave as graph density increases. Denser graphs benefit more consistently from PhyloEvolve’s optimizations, yielding stable performance gains relative to the baseline. This improvement arises from the system’s ability to reorganize sparse-matrix operations, reduce kernel-launch fragmentation, and exploit shared-memory tiling in repeated neighborhood aggregations. As density grows, the workload becomes more regular and amenable to GPU parallelization, allowing PhyloEvolve to generate efficient execution patterns.}
\label{fig:subim2}
\end{subfigure}
\caption{Performance improvements achieved by PhyloEvolve on the GraphWave algorithm across varying graph sizes (a) and densities (b). GraphWave relies heavily on spectral graph computations, including Laplacian eigendecomposition and heat-kernel-based wave propagation. PhyloEvolve autonomously discovers GPU-efficient kernel variants that outperform the baseline implementation, especially on larger and denser graphs. These optimizations arise from trajectory-conditioned learning rather than random mutation, enabling the system to restructure sparse-linear-algebra operations, reduce kernel launch overhead, and exploit batching opportunities. The results highlight PhyloEvolve’s ability to adapt to different graph regimes and improve throughput in spectral embedding workflows.}
\label{fig:image2}
\end{figure}

\section{Discussion}
\tool reframes GPU algorithm optimization as a trajectory-centric in-context reinforcement learning problem rather than a mutation-driven evolutionary search. Across diverse workloads—including stiff PDE solvers, manifold-learning pipelines, and spectral graph algorithms—the system consistently discovers performance-enhancing transformations that restructure memory layouts, reduce redundant operations, and increase GPU utilization. These improvements emerge from reusable optimization knowledge encoded in evolving trajectories rather than from hand-tuned heuristics or population-level mutation statistics. By elevating trajectories to first-class learning signals, \tool captures both high-level optimization strategies and fine-grained edit patterns that can be reused and recombined across tasks. A central outcome of our experiments is that trajectory-conditioned optimization meaningfully reduces redundant exploration. Traditional evolutionary algorithms repeatedly generate uncontextualized mutations and rely on large populations to smooth out stochasticity. In contrast, \tool uses prior successful trajectories from across the phylogenetic forest—even those originating in unrelated computational regimes—to guide refinement on new tasks. This mechanism leads to faster convergence, more stable improvements, and more consistent performance across heterogeneous workloads. In several cases, the system identifies optimization motifs—such as kernel fusion patterns, shared-memory tiling, and batching strategies—that generalize across algorithmic families, suggesting the emergence of a cross-domain meta-optimization effect.

This work also highlights several important limitations that point toward promising directions for future research. First, \tool currently relies on empirical evaluation alone when proposing refinements, without incorporating hardware-aware cost models or analytical performance estimates. This makes the system robust and general but can be inefficient for computationally heavy workloads. Integrating analytic or learned cost models could improve sample efficiency and reduce unnecessary evaluations. Second, our experiments focus on NVIDIA GPUs; although the architecture is designed to be backend-agnostic, evaluating portability across AMD, Intel, or cloud accelerators is necessary to establish broader applicability. Likewise, scaling beyond single-GPU execution to multi-GPU or distributed settings remains open, as communication and synchronization introduce optimization challenges not yet addressed in our framework. Another limitation is that while the phylogenetic forest encourages diversity, variation in LLM-generated modifications can still lead to occasional low-quality proposals. Our ModifyAgent mitigates this through filtering and compilation checks, but more structured constraint mechanisms—or learned validity priors—may further improve reliability. Additionally, although \tool demonstrates cross-task generalization empirically, we do not yet provide a theoretical understanding of when trajectory reuse yields consistent improvement operators or under what assumptions in-context learning induces stable optimization behavior. Formal analysis of these dynamics would deepen the conceptual foundation of ICRL-driven optimization systems. Ensuring correctness is another practical concern for scientific workloads. While we verify numerical accuracy empirically, future extensions could incorporate lightweight formal verification, interval arithmetic, or symbolic reasoning to guarantee correctness without sacrificing flexibility. Finally, the system generates large numbers of optimization trajectories; techniques for trajectory compression, clustering, or extraction of recurrent “optimization motifs” could enable higher-level primitives analogous to compiler passes, making the system more interpretable and scalable.

Overall, \tool demonstrates that structured trajectory reuse can serve as a powerful alternative to mutation-centric evolutionary search, enabling consistent, cross-domain improvements in GPU algorithm optimization. At the same time, addressing the limitations outlined above—incorporating cost models, extending hardware coverage, improving proposal validity, and formalizing theoretical underpinnings—will be essential for scaling this approach to broader classes of computational workloads and multi-accelerator systems.

\section{Conclusion}
We presented \tool, a system that reframes GPU algorithm optimization as a trajectory-driven in-context reinforcement learning process. Rather than relying on mutation-centric evolutionary procedures or static LLM refactoring, \tool captures and reuses structured optimization trajectories through a phylogenetic forest, enabling informed, context-aware refinements across heterogeneous computational workloads. By treating trajectories as first-class learning signals, the system autonomously identifies optimization strategies—from memory-layout adjustments to kernel fusion and batching heuristics—that generalize across algorithmic families without retraining.

Our experiments demonstrate that \tool consistently produces substantial performance gains for workloads ranging from stiff PDE solvers to manifold learning and spectral graph computations. These improvements arise from trajectory-conditioned experience reuse, elite-pool cross-lineage transfer, and multi-agent coordination, collectively reducing redundant exploration and accelerating convergence. Importantly, the system uncovers domain-transferable optimization motifs that emerge naturally from in-context learning rather than manual heuristics or explicit rules. While the results are encouraging, this work also exposes directions for further development, including incorporation of hardware-aware priors, extension to distributed and heterogeneous accelerator environments, improved filtering of LLM-generated proposals, and deeper theoretical understanding of trajectory-induced improvement dynamics. Addressing these challenges will be essential for scaling \tool to more complex scientific computing pipelines and multi-device architectures.

Overall, \tool demonstrates that trajectory-centric ICRL provides a promising foundation for automated algorithm optimization. By unifying structured memory, learning-based refinement, and cross-task generalization, the framework opens a path toward next-generation LLM-guided meta-optimization systems capable of evolving high-performance computational kernels across a broad range of scientific and engineering domains.

\newpage

\bibliography{reference}
\bibliographystyle{unsrt}

\appendix
% You may include other additional sections here.

%\input{tex/appendix/forest_full}

%\bibliographystyle{unsrtnat}
%\bibliography{references}  %%% Uncomment this line and comment out the ``thebibliography'' section below to use the external .bib file (using bibtex) .

%%% Uncomment this section and comment out the \bibliography{references} line above to use inline references.
% \begin{thebibliography}{1}

% 	\bibitem{kour2014real}
% 	George Kour and Raid Saabne.
% 	\newblock Real-time segmentation of on-line handwritten arabic script.
% 	\newblock In {\em Frontiers in Handwriting Recognition (ICFHR), 2014 14th
% 			International Conference on}, pages 417--422. IEEE, 2014.

% 	\bibitem{kour2014fast}
% 	George Kour and Raid Saabne.
% 	\newblock Fast classification of handwritten on-line arabic characters.
% 	\newblock In {\em Soft Computing and Pattern Recognition (SoCPaR), 2014 6th
% 			International Conference of}, pages 312--318. IEEE, 2014.

% 	\bibitem{hadash2018estimate}
% 	Guy Hadash, Einat Kermany, Boaz Carmeli, Ofer Lavi, George Kour, and Alon
% 	Jacovi.
% 	\newblock Estimate and replace: A novel approach to integrating deep neural
% 	networks with existing applications.
% 	\newblock {\em arXiv preprint arXiv:1804.09028}, 2018.

% \end{thebibliography}

\end{document}